\documentclass[11pt]{article}

\usepackage[final]{acl}
\usepackage{hyperref}
\usepackage{times}
\usepackage{latexsym}
\usepackage{booktabs,longtable}
\usepackage{subcaption}

\usepackage[T1]{fontenc}

\usepackage[utf8]{inputenc}

\usepackage{microtype}

\usepackage{inconsolata}
\usepackage{graphicx}

\usepackage{amsmath, amssymb}

\title{Prompt Robustness Is Task-Dependent: Comparing Objective and Belief-Style Questions in LLM Evaluation}

\author{
\normalfont{
\parbox{\linewidth}{
\centering
Sadia Kamal$^{\dagger}$, 
Arefa Patwary$^{\ddagger}$,
Anthony Marchiafava$^{\dagger}$,  
Sagnik Ray Choudhury$^{\ddagger}$,
Atriya Sen$^{\dagger}$\\
\parbox{\linewidth}{
\centering
  $^{\dagger}$Oklahoma State University,
  $^{\ddagger}$University of North Texas \\
  \texttt{\{sadia.kamal,anmarch,atriya.sen\}@okstate.edu,  arefapatwary@my.unt.edu,sagnik.raychoudhury@unt.edu}
  }
  }
}
}

\begin{document}
\maketitle



\begin{abstract}
Survey-style evaluations of large language models often treat a prompted response as a measure of a model's values or beliefs. This assumption is particularly fragile when responses are read as evidence of political values, social attitudes, or beliefs. We ask whether prompt robustness differs between objective questions with fixed answers and subjective questions that ask for opinions or values. We evaluate four instruction-tuned model families on three objective datasets (MMLU, ARC, and CulturalBench) and three subjective datasets (Political Compass Test, ValueBench, and World Values Survey). For each question/statement, we apply multiple types of prompt changes, such as variations in wording, framing, and format, and measure whether the model gives the same answer across variants. Using a binomial generalized estimating equation, we find significant effects of model, dataset, prompt category, and their interactions. The dataset type effect is also significant, and the interaction between dataset type and prompt category is large. These results show that prompt robustness depends on the question type, the prompt change, and the model. 

\end{abstract}

\section{Introduction}

Most large language model evaluations rest on a fragile assumption that one prompt gives a stable and meaningful measure of model behavior. Prior work shows this is often not the case, small changes in prompt format, wording, and answer presentation can change model behavior \citep{sclar2024quantifying,chatterjee2024posix,ismithdeen2025promptception}. If a model changes its answer when the task meaning stays the same, then the evaluation is measuring both task ability and prompt sensitivity.

This is a critical concern for survey-style tests. Recent work uses political and value surveys to infer what LLMs ``believe'' or which human groups they resemble. But survey responses from LLMs are unstable under ordering, labeling, forced-choice wording, and framing changes \citep{dominguez2024questioning,rottger2024political,rupprecht2025prompt}. A model's answer can be shaped by the way the prompt asks the question.

This raises a question about whether prompt sensitivity behaves the same way for different types of questions. We study this problem by comparing two kinds of questions. Type-I questions consist of objective, multiple-choice items with a single correct answer. Type-II questions are subjective survey items that ask for opinions, values, or degrees of agreement. When a prompt is reworded without altering its underlying meaning, a robust model is expected to produce the same answer across both versions. The same applies for objective questions as well. For subjective questions, however, models may interpret minor changes in wording or response options as cues about how to respond. As a result, the model's answer can shift even when the survey item itself has not changed in substance.

We ask three research questions: \textbf{RQ1:} Does response consistency differ between objective and subjective question types? \textbf{RQ2:} Do objective and subjective questions show different sensitivity patterns across prompt categories? \textbf{RQ3:} Is prompt robustness a model-level property, or does it also depend on dataset and prompt category?

Our study makes three contributions. First, we curate a broad perturbation set from prior work on prompt sensitivity, multiple-choice formatting, survey response bias, and value measurement. Second, we give a unified robustness test across objective and subjective datasets. Third, we use clustered statistical tests to show that prompt sensitivity depends on model, dataset, dataset type, and perturbation category.  We find that subjective questions are less stable under prompt variation than objective questions. The gap is not uniform across perturbation categories: option-order changes produce the largest consistency drop, while lexical and logically equivalent perturbations are comparatively stable. Statistical analysis shows that robustness depends jointly on model, dataset, dataset type, and prompt category. Our findings show that LLM prompt sensitivity varies systematically with question type and should be evaluated across perturbation families rather than through a single prompt.

\section{Related Work}

\paragraph{Prompt sensitivity in LLM evaluation.}
Sclar et al.\citep{sclar2024quantifying} show that prompt formatting can produce large accuracy differences and argue that evaluation should report variation across plausible prompt formats, not only one prompt . POSIX\citep{chatterjee2024posix} measures prompt sensitivity through likelihood change under intent-preserving prompt variants . Promptception\citep{ismithdeen2025promptception} builds a broad perturbation framework for multiple-choice multimodal evaluation, including phrasing, formatting, position, and option presentation . These works show that prompt design is part of the evaluation, not a neutral wrapper around it.

\paragraph{LLM survey responses.}
Survey-style evaluation has been used to study model values, politics, and social attitudes. But recent studies warn that such responses can be strongly affected by answer ordering, labels, forced-choice formats, and paraphrases \citep{dominguez2024questioning,rottger2024political,wright2024llmtropes,rupprecht2025prompt}. This makes political and value scores difficult to interpret as stable model traits. Our work adds a direct comparison with objective multiple-choice datasets, which helps separate general prompt noise from instability that is specific to subjective questions.

\section{Experimental Design}

\subsection{Datasets and Models}

We use six datasets categorized into two types. The Type-I group contains objective multiple-choice questions: MMLU \citep{hendrycks2021measuring}, ARC \citep{clark2018arc}, and CulturalBench-Easy \citep{chiu2024culturalbench}. The Type-II group contains subjective or opinion-based statements: the Political Compass Test, ValueBench \citep{ren2024valuebench}, and World Values Survey items \citep{haerpfer2022worldvalues}. We use CulturalBench-Easy as a Type-I dataset because it tests cultural knowledge through multiple-choice questions with fixed answer keys, rather than asking for subjective preferences or beliefs. We evaluate four instruction-tuned model families: Gemma, Llama, Mistral, and Qwen.

\subsection{Perturbation Curation}

We conducted an extensive literature search across three lines of work: prompt robustness, multiple-choice evaluation, and LLM survey measurement. From this search, we selected perturbations only if they met the following criteria. First, the perturbation had to preserve the intended task for both factual and survey-style items. Second, it had to test a known source of instability, such as wording, option labels, option order, or formatting. Table~\ref{tab:perturbation-taxonomy} gives the final taxonomy.

\begin{table*}[t]
\centering
\small
\begin{tabular}{p{0.16\textwidth}p{0.27\textwidth}p{0.29\textwidth}p{0.18\textwidth}}
\toprule
Category & Operation & What it tests & Literature \\
\midrule
Paraphrase & Rewrite the instruction with the same task intent. & Sensitivity to semantically similar wording. & Prompt robustness and survey paraphrase effects \citep{sclar2024quantifying,chatterjee2024posix,rottger2024political}. \\
Spelling noise & Add small spelling mistakes while keeping the item readable. & Robustness to surface noise and user typos. & Prompt perturbation and input-noise evaluation \citep{ismithdeen2025promptception}. \\
Lexical substitution & Replace selected words with near synonyms. & Dependence on word choice when the intended meaning is stable. & Intent-preserving prompt variation \citep{chatterjee2024posix}. \\
Logical equivalent & Use truth-preserving instruction forms. & Whether logically equivalent task descriptions keep the same answer. & Controlled prompt rewriting \citep{sclar2024quantifying,chatterjee2024posix}. \\
Label substitution & Replace A, B, C, D with numbers, Roman numerals, rare Unicode symbols, or regular symbols. & Dependence on answer labels rather than option content. & Multiple-choice option sensitivity and survey response artifacts \citep{ismithdeen2025promptception,rupprecht2025prompt}. \\
Format variation & Change separators, spacing, casing, wrappers, joiners, and field layout. & Sensitivity to prompt layout rather than semantic content. & Prompt formatting sensitivity \citep{sclar2024quantifying,ismithdeen2025promptception}. \\
Option shuffling & Reverse, rotate, or swap option order while tracking the correct answer when one exists. & Position bias, recency bias, and order effects. & Survey and multiple-choice ordering effects \citep{dominguez2024questioning,rupprecht2025prompt}. \\
\bottomrule
\end{tabular}
\caption{Prompt perturbation taxonomy. The final categories were selected from prior work and then adapted so the same robustness design could be used for objective and subjective datasets.}
\label{tab:perturbation-taxonomy}
\end{table*}

\subsection{Prompt Generation and Answer Normalization}

For each item, we create prompt variants that preserve the intended task. All prompts use the same broad instruction: the model must select exactly one labeled option. For subjective datasets, the labels map to agreement or value options. For objective datasets, the labels map to answer choices. We keep this common format so that the Type-I and Type-II groups differ mainly in question content, not in answer extraction. All runs use deterministic decoding with temperature 0 so that answer changes come from prompt variation rather than sampling noise.

The perturbation set is designed to separate three failure sources. The first source is semantic framing. Paraphrase, lexical substitution, spelling noise and logical equivalence ask whether two prompts with the same intended task produce the same answer. The second source is surface form. Format variation test whether the model is distracted by small changes that should not alter the task. The third source is answer presentation. Label substitution and option shuffling test whether the model follows the content of the options or instead follows labels and positions. This separation matters because a single consistency score can hide different causes of instability.

We use the same perturbation logic across objective and subjective datasets, but the interpretation differs. On objective datasets, the answer key gives an external standard. If the model changes from one option to another under a meaning-preserving change, this is a robustness failure. On subjective datasets, there is no single correct option. Therefore, we do not call one answer right and another wrong. We instead ask whether the model keeps its own answer stable when the survey item is presented in a controlled new form.

We normalize outputs before computing consistency. For objective datasets, raw outputs are mapped to canonical A, B, C, or D labels. For subjective datasets, agreement labels are also mapped to canonical letters. For substitution runs, symbols are mapped back to their original A/B/C/D meanings. This prevents a numeric answer such as ``1'' and a letter answer such as ``A'' from being counted as inconsistent when they refer to the same option.

\subsection{Consistency Metric and Statistical Model}

Our primary evaluation metric is answer consistency. It measures whether a model gives the same answer across prompt variants. Throughout this section, we will refer a single question from a dataset as an item. For each item, model, and prompt category, we compute a consistency ratio:
\begin{equation}
    C_i = \frac{\max_y n_i(y)}{N_i},
\end{equation}
where $n_i(y)$ is the number of times answer $y$ appears for item $i$, and $N_i$ is the number of valid prompt variants for that item. A value of 1 means all variants gave the same answer. Lower values mean the prompt category changed the answer.

We fit binomial generalized estimating equation (GEE) models \citep{liang1986longitudinal}. GEE is useful here because each item appears many times under related prompt variants, so rows are not independent. The response is whether a prompt variant matches the majority answer for that item. We test main effects for model, dataset, dataset type, and prompt category, plus interactions between dataset or dataset type and prompt category, and between model and prompt category. We report Wald $\chi^2$ tests.

\begin{figure*}[t]
\centering
\begin{subfigure}[t]{0.49\textwidth}
    \centering
    \includegraphics[width=\linewidth]{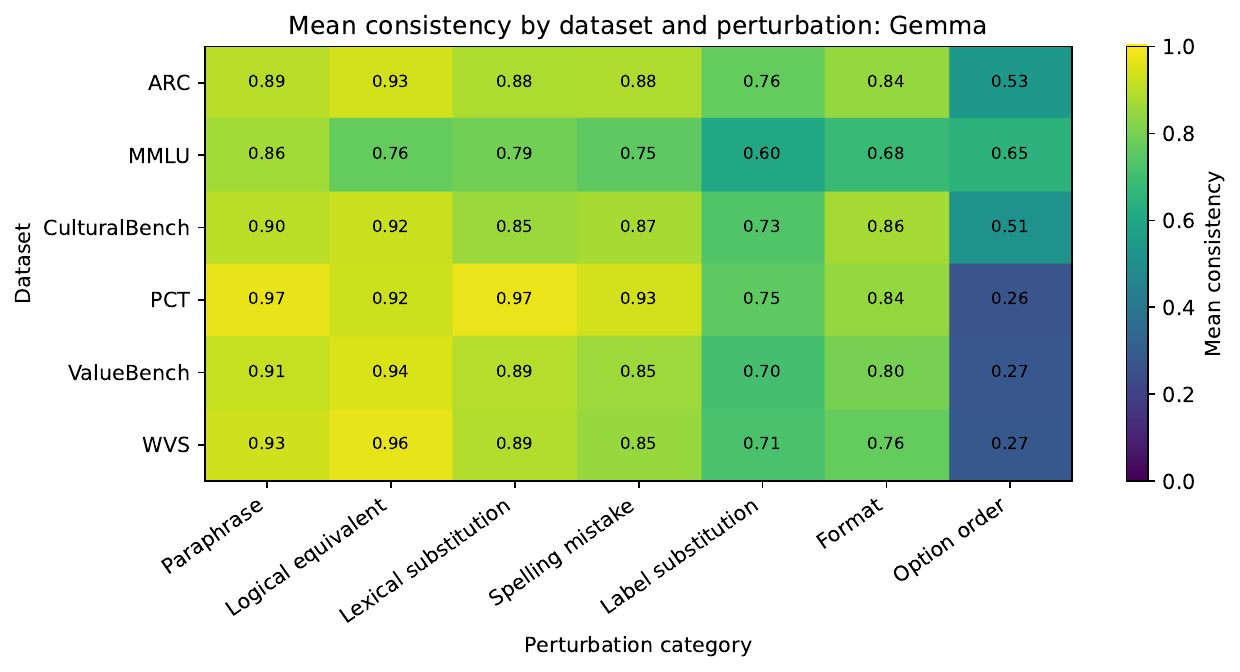}
    \caption{Gemma}
\end{subfigure}
\hfill
\begin{subfigure}[t]{0.49\textwidth}
    \centering
    \includegraphics[width=\linewidth]{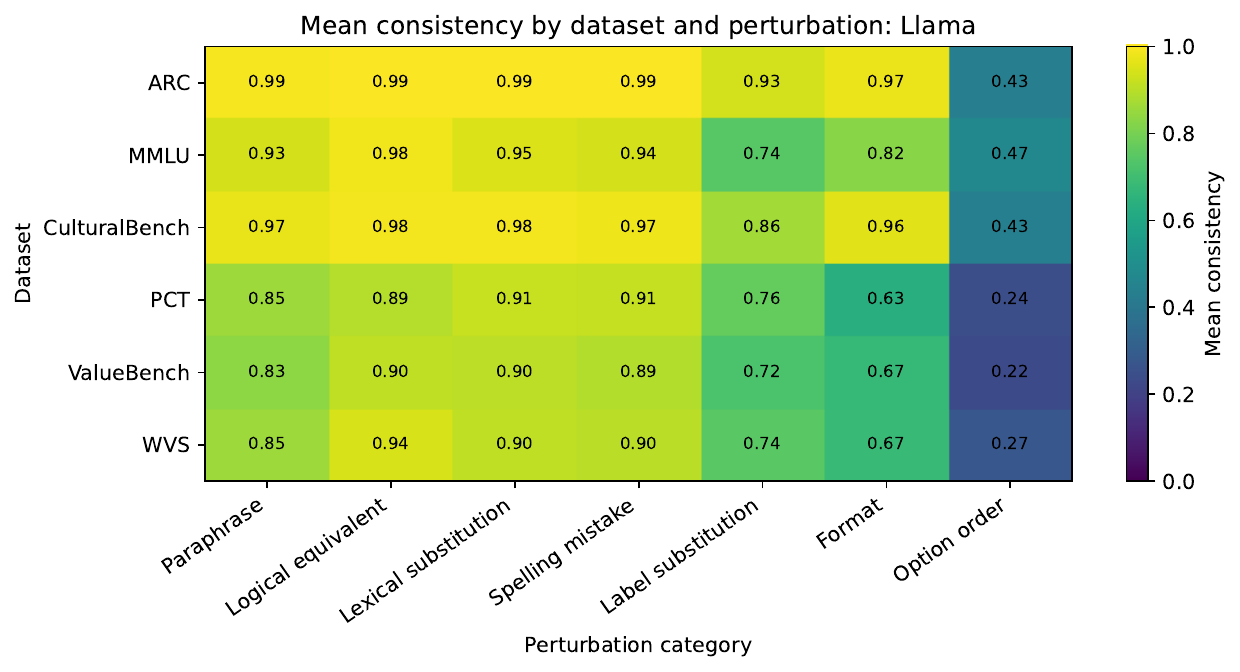}
    \caption{Llama}
\end{subfigure}

\vspace{0.6em}

\begin{subfigure}[t]{0.49\textwidth}
    \centering
    \includegraphics[width=\linewidth]{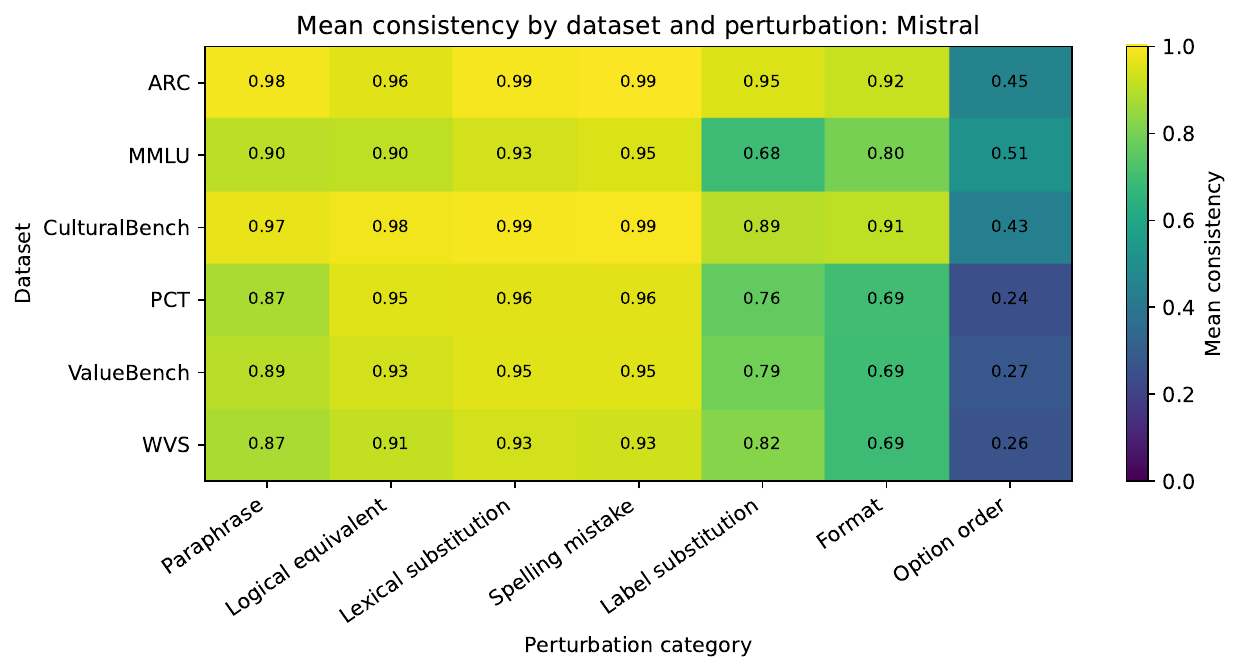}
    \caption{Mistral}
\end{subfigure}
\hfill
\begin{subfigure}[t]{0.49\textwidth}
    \centering
    \includegraphics[width=\linewidth]{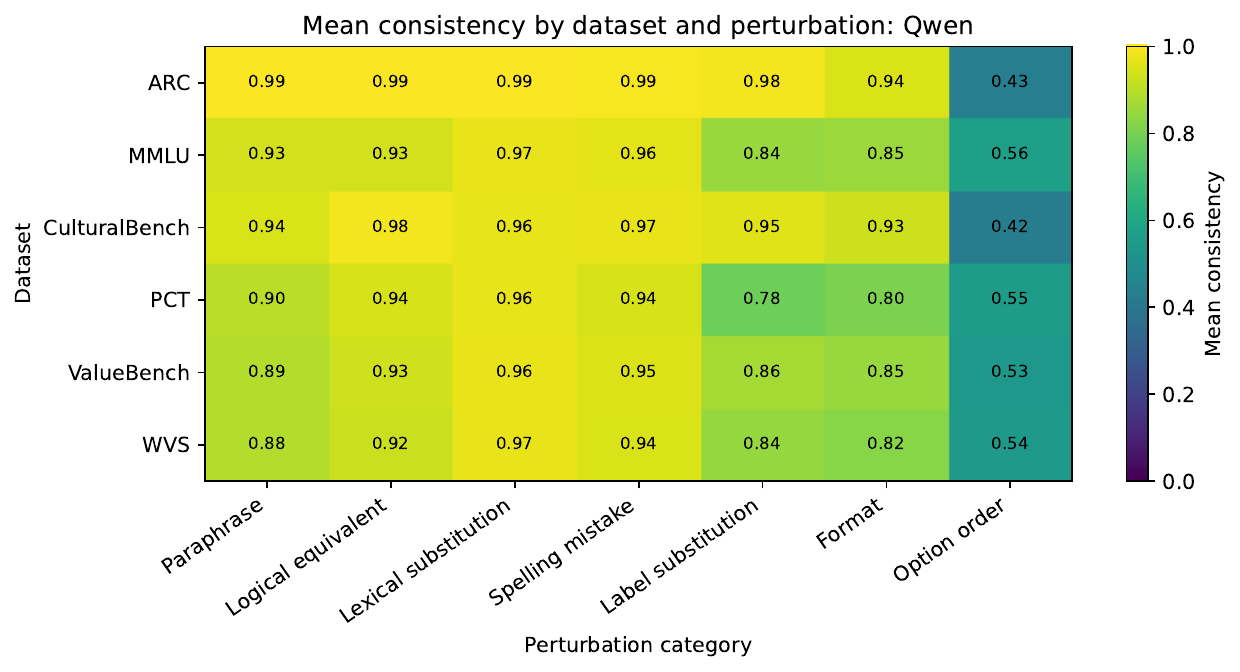}
    \caption{Qwen}
\end{subfigure}

\caption{Dataset-level mean consistency across perturbation categories for all four models.}
\label{fig:modelwise-heatmap-panel}
\end{figure*}

\section{Results}

\subsection{Subjective Questions Are More Affected Than Objective Questions}

Figure~\ref{fig:modelwise-heatmap-panel} shows that, across the four models, subjective datasets usually have lower consistency than objective datasets under most perturbation categories. Averaged over all perturbations, Type-I objective datasets have a mean consistency of 0.849, while Type-II subjective datasets have a mean consistency of 0.787. Equivalently, instability rises from 0.151 for objective datasets to 0.213 for subjective datasets.

Subjective tasks are consistently less stable than objective ones across all four models, which answers RQ1. The gap is persistent across Gemma, Llama, Mistral, and Qwen models, although the magnitude differs by model. It reflects a systematic difference between question types.

Table~\ref{tab:type-perturbation-means} summarizes the marginal consistency scores by dataset type and perturbation category. The Type-I versus Type-II gap is largest under option order. For option-order perturbations, Type-I consistency is 0.485, while Type-II consistency is only 0.328. This means subjective datasets add 15.7 percentage points of extra instability beyond the already large option-order effect in objective datasets. The next largest gaps appear under format variation and label substitution. The smallest gaps appear under lexical substitution and logical equivalence, where both dataset types remain relatively stable.

\subsection{Option Order Is the Strongest Source of Instability}

Having established that subjective questions are less robust overall, we next ask which perturbation category drives the largest instability. Across all four models and all six datasets, option order perturbation category affects the most. Averaged over models and datasets, option-order perturbations reduce mean consistency to 0.407. This is much lower than label substitution, 0.799, and format variation, 0.808. Semantic and near-semantic perturbations are more stable: paraphrase reaches 0.913, spelling noise reaches 0.927, logical equivalence reaches 0.935, and lexical substitution reaches 0.936.

This pattern is visible in each model in Figure~\ref{fig:modelwise-heatmap-panel}. For Gemma, Llama, and Mistral, option order produces the lowest consistency values for the subjective datasets, particularly PCT, ValueBench, and WVS. Qwen is more robust overall, but option order still produces its clearest drop compared with semantic perturbations. Thus, the main finding is not that models are equally fragile to all prompt changes. Instead, answer-presentation perturbations, particularly option order, create the largest instability.

\begin{table}[t]
\centering
\small
\begin{tabular}{lrrr}
\toprule
Perturbation & Type-I & Type-II & Gap \\
\midrule
Option order & 0.485 & 0.328 & 0.157 \\
Format & 0.873 & 0.743 & 0.130 \\
Label substitution & 0.827 & 0.771 & 0.057 \\
Paraphrase & 0.938 & 0.888 & 0.050 \\
Spelling noise & 0.938 & 0.917 & 0.021 \\
Logical equivalent & 0.943 & 0.928 & 0.015 \\
Lexical substitution & 0.938 & 0.933 & 0.005 \\
\bottomrule
\end{tabular}
\caption{Mean consistency by dataset type and perturbation category. Gap is computed as Type-I consistency minus Type-II consistency, so larger values indicate greater subjective-task instability.}
\label{tab:type-perturbation-means}
\end{table}

This addresses RQ2, objective and subjective questions do not merely differ in overall robustness. They differ in the kinds of prompt changes that affect them most. The subjective-task gap is largest when the prompt changes how answers are presented, rather than when it only changes lexical wording.

\subsection{Dataset and Prompt Category Interact Strongly}

Table~\ref{tab:global-dataset} demonstrates the global GEE model with dataset as a six-level factor. Model, dataset, and prompt category are all significant predictors of consistency. More importantly, the dataset by prompt category interaction is large and significant, $\chi^2=587.885$, $p=9.52\times10^{-105}$. This confirms the pattern in Figure~\ref{fig:modelwise-heatmap-panel}: perturbation categories do not affect all datasets in the same way. The model by prompt category interaction is also significant, $\chi^2=616.748$, $p=2.47\times10^{-119}$, showing that models differ in which perturbations they are most sensitive to.

\begin{table}[t]
\centering
\small
\begin{tabular}{lrrr}
\toprule
Effect & $\chi^2$ & df & $p$ \\
\midrule
Model & 173.124 & 3 & $2.69\times10^{-37}$ \\
Dataset & 463.831 & 5 & $5.10\times10^{-98}$ \\
Prompt category & 523.719 & 6 & $6.52\times10^{-110}$ \\
Dataset $\times$ prompt & 587.885 & 30 & $9.52\times10^{-105}$ \\
Model $\times$ prompt & 616.748 & 18 & $2.47\times10^{-119}$ \\
\bottomrule
\end{tabular}
\caption{Global Wald $\chi^2$ tests with dataset as a six-level factor.}
\label{tab:global-dataset}
\end{table}

\subsection{Dataset Type Explains a Systematic Robustness Gap}

Table~\ref{tab:global-type} divides datasets into Type-I and Type-II groups. Dataset type remains highly significant, $\chi^2=256.026$, $p=1.26\times10^{-57}$. The interaction between dataset type and prompt category is also significant, $\chi^2=175.435$, $p=3.16\times10^{-35}$. This interaction shows that subjective datasets are not merely lower in consistency on average. They are differently affected by different perturbation categories.

\begin{table}[t]
\centering
\small
\begin{tabular}{lrrr}
\toprule
Effect & $\chi^2$ & df & $p$ \\
\midrule
Model & 171.898 & 3 & $4.95\times10^{-37}$ \\
Dataset type & 256.026 & 1 & $1.26\times10^{-57}$ \\
Prompt category & 709.099 & 6 & $6.64\times10^{-150}$ \\
Type $\times$ prompt & 175.435 & 6 & $3.16\times10^{-35}$ \\
Model $\times$ prompt & 617.207 & 18 & $1.98\times10^{-119}$ \\
\bottomrule
\end{tabular}
\caption{Global Wald $\chi^2$ tests with dataset type as Type-I versus Type-II.}
\label{tab:global-type}
\end{table}

The results underscores that single-prompt survey measurements are notably fragile. For objective multiple-choice questions, there is an external target answer that can constrain the model. For subjective survey items, small changes in wording, formatting, labels, or option order can shift the response without any external gold answer. This makes a single response difficult to interpret as a stable value or belief.

\subsection{All Datasets Show Prompt Effects}

Table~\ref{tab:within} reports the prompt-category effect from within-dataset GEE models. Prompt category is significant for all six datasets. The largest effects appear in the subjective datasets: WVS, $\chi^2=1146.897$, ValueBench, $\chi^2=993.681$, and PCT, $\chi^2=757.436$. Objective datasets also show significant effects, but the magnitudes are smaller: ARC, $\chi^2=191.625$, MMLU, $\chi^2=96.950$, and CulturalBench, $\chi^2=279.848$.

\begin{table}[t]
\centering
\small
\begin{tabular}{lrrr}
\toprule
Dataset & $\chi^2$ & df & $p$ \\
\midrule
ARC & 191.625 & 6 & $1.15\times10^{-38}$ \\
MMLU & 96.950 & 6 & $1.09\times10^{-18}$ \\
CulturalBench & 279.848 & 6 & $1.69\times10^{-57}$ \\
PCT & 757.436 & 6 & $2.41\times10^{-160}$ \\
ValueBench & 993.681 & 6 & $2.08\times10^{-211}$ \\
WVS & 1146.897 & 6 & $1.49\times10^{-244}$ \\
\bottomrule
\end{tabular}
\caption{Prompt-category effects from within-dataset GEE models.}
\label{tab:within}
\end{table}

This result rules out a narrow explanation in which prompt sensitivity is only a survey problem. Prompt variation affects all datasets. However, the effect is more consequential for subjective datasets because a shifted answer can change the inferred political position, social value, or belief attributed to the model.

\subsection{Robustness Varies Across Models, Datasets, and Perturbations}

The model-wise heatmaps in Figure~\ref{fig:modelwise-heatmap-panel} show that robustness is not a single model trait. A model can be stable under paraphrase or lexical substitution but unstable under option order. The same model can also behave differently across objective and subjective datasets. For example, Qwen is generally more stable than the other models under option order, but it still shows a visible drop compared with semantic perturbations. Llama and Mistral are highly stable on several objective cells but much less stable on subjective option-order cells.

The model by prompt category interactions in Tables~\ref{tab:global-dataset} and~\ref{tab:global-type} confirm this observation statistically. Appendix Table~\ref{tab:model-dataset-prompt-gee} reports the full model by dataset prompt-effect table. Every model-dataset bin shows a significant prompt-category effect. Therefore, robustness should be reported as a relation among model, dataset, and perturbation category, not as a single global score.

This matters for model comparison. A model that appears robust on MMLU paraphrases may not be robust on WVS option order or PCT label substitution. A single average robustness score can therefore hide important failures. For survey-style evaluations, the most important warning is that a single prompted answer should not be treated as a stable measurement of model values or beliefs.

\section{Discussion}

These results have a direct implication for LLM survey studies. A single answer to a political or value question should not be treated as a stable model belief unless it survives prompt variation. This does not mean survey-style evaluation is useless. It means that survey scores need robustness checks. If two meaning-preserving prompts produce different answers, the result cannot be attributed to the model alone.
The objective-subjective comparison sheds light on why this happens. Objective questions have fixed answers, so a change in answer usually signals task failure or prompt sensitivity. Subjective questions are different. A model may use wording, label choice, or option order as evidence about how it should respond. In that case, prompt variation is not only noise. It is also a way to reveal how fragile the measured stance is.
Researchers using survey-style evaluations should therefore report consistency across prompt variants, not only a final accuracy or survey score. Semantic perturbations should be separated from surface perturbations, since they reveal different failure modes. Interactions with dataset type should also be reported. 

Beyond measuring consistency, the perturbation taxonomy can identify where instability comes from. Failures under spelling or format changes suggest brittle instruction following. Failures under label substitution or option shuffling suggest answer-position bias or label-token bias. Failures under paraphrase or logical equivalence are the most consequential for survey interpretation, because the measured stance changes even when the item meaning is intended to stay fixed. These failure types should not be collapsed into a single error category.

\section{Limitations}

This study uses deterministic decoding and forced-choice answers. This helps isolate prompt effects, but it does not cover open-ended survey responses. The model names are treated at the family level, so future work should report exact checkpoint versions and parameter sizes. Finally, subjective datasets are not interchangeable. PCT, ValueBench, and WVS differ in wording, scale design, and social meaning.

\section{Conclusion}

We compared prompt robustness across objective and subjective questions using six datasets and four model families. Consistency depends strongly on model, dataset, prompt category, and dataset type. Subjective survey tasks show substantially larger prompt-category effects than objective tasks, and these effects vary by both the type of prompt change and the model being tested. Characterizing model robustness therefore requires systematic variation across prompt forms. For value and belief evaluation, robustness to prompt variation should be a standard part of the evaluation design.

\bibliography{custom}


\section{Appendix}
\label{sec:appendix}
\section{Full Model by Dataset Prompt Effects}
Table~\ref{tab:model-dataset-prompt-gee} reports the within-cell prompt-category effect for each model and dataset pair. Each row comes from a separate binomial GEE model fit within one model-dataset cell. The table shows that prompt category has a significant effect in every model-dataset combination, which supports the main claim that prompt robustness is not a single global model property. Instead, it varies with the model, dataset, and perturbation category.
\begin{table}[h]
\centering
\scriptsize
\begin{tabular}{llrrr}
\toprule
Model & Dataset & $\chi^2$ & df & $p$ \\
\midrule
Gemma & ARC & 191.625 & 6 & $1.15\times10^{-38}$ \\
Gemma & MMLU & 96.950 & 6 & $1.09\times10^{-18}$ \\
Gemma & CulturalBench & 279.848 & 6 & $1.69\times10^{-57}$ \\
Gemma & PCT & 757.436 & 6 & $2.41\times10^{-160}$ \\
Gemma & ValueBench & 993.681 & 6 & $2.08\times10^{-211}$ \\
Gemma & WVS & 1146.897 & 6 & $1.49\times10^{-244}$ \\
\midrule
Llama & ARC & 281.043 & 6 & $9.39\times10^{-58}$ \\
Llama & MMLU & 333.402 & 6 & $5.63\times10^{-69}$ \\
Llama & CulturalBench & 331.394 & 6 & $1.52\times10^{-68}$ \\
Llama & PCT & 1367.245 & 6 & $3.00\times10^{-292}$ \\
Llama & ValueBench & 1765.928 & 6 & $<10^{-300}$ \\
Llama & WVS & 1632.529 & 6 & $<10^{-300}$ \\
\midrule
Mistral & ARC & 504.350 & 6 & $9.72\times10^{-106}$ \\
Mistral & MMLU & 314.800 & 6 & $5.50\times10^{-65}$ \\
Mistral & CulturalBench & 450.267 & 6 & $4.30\times10^{-94}$ \\
Mistral & PCT & 1276.720 & 6 & $1.19\times10^{-272}$ \\
Mistral & ValueBench & 1658.887 & 6 & $<10^{-300}$ \\
Mistral & WVS & 2245.270 & 6 & $<10^{-300}$ \\
\midrule
Qwen & ARC & 1006.535 & 6 & $3.45\times10^{-214}$ \\
Qwen & MMLU & 220.672 & 6 & $7.48\times10^{-45}$ \\
Qwen & CulturalBench & 407.518 & 6 & $6.76\times10^{-85}$ \\
Qwen & PCT & 172.618 & 6 & $1.25\times10^{-34}$ \\
Qwen & ValueBench & 502.756 & 6 & $2.14\times10^{-105}$ \\
Qwen & WVS & 587.695 & 6 & $1.05\times10^{-123}$ \\
\bottomrule
\end{tabular}
\caption{Prompt-category effects within each model-dataset binomial GEE model. This is the full version of the model by dataset result table.}
\label{tab:model-dataset-prompt-gee}
\end{table}

\clearpage

\end{document}